# Convolutional Sparse Coding for High Dynamic Range Imaging

Ana Serrano[1]      Felix Heide[2]      Diego Gutierrez[1]      Gordon Wetzstein[2]      Belen Masia[1,3]

[1] Universidad de Zaragoza      [2] Stanford University      [3] MPI Informatik

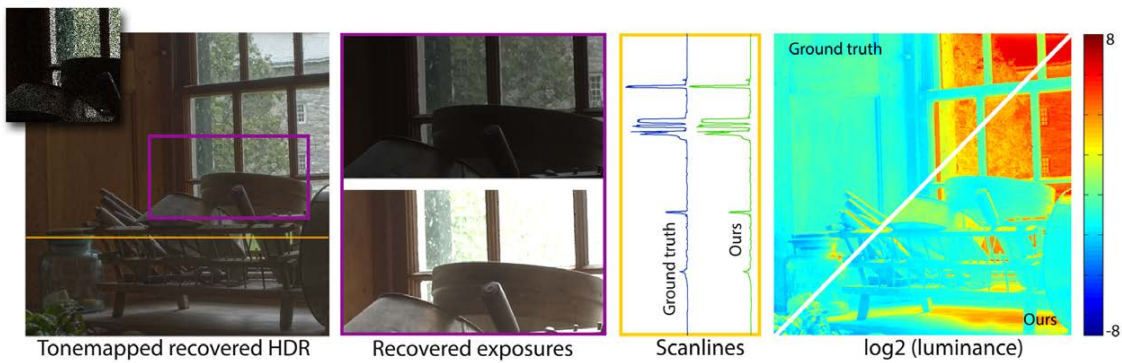

**Figure 1:** *High dynamic range image (HDRI) recovered from a single, coded, 8-bit low dynamic range (LDR) image using the proposed sparse reconstruction method. Left: HDR image recovered with our framework, tonemapped for display purposes. The inset shows a cropped region of the coded LDR image used as input to the reconstruction algorithm. Center left: Close-up of two exposures of the reconstructed HDR image showing the ability of our method to reconstruct an extended dynamic range. Center right: normalized luminance plots of the marked scanline (yellow line, rotated by 90°) for the reconstructed image (green curve) and the ground truth image (blue curve). Right: false color image of the reconstructed HDR scene (scale is in stops), showing the extremely large dynamic range that the original scene had and our technique is able recover.*

**Abstract**

*Current HDR acquisition techniques are based on either (i) fusing multibracketed, low dynamic range (LDR) images, (ii) modifying existing hardware and capturing different exposures simultaneously with multiple sensors, or (iii) reconstructing a single image with spatially-varying pixel exposures. In this paper, we propose a novel algorithm to recover high-quality HDRI images from a single, coded exposure. The proposed reconstruction method builds on recently-introduced ideas of convolutional sparse coding (CSC); this paper demonstrates how to make CSC practical for HDR imaging. We demonstrate that the proposed algorithm achieves higher-quality reconstructions than alternative methods, we evaluate optical coding schemes, analyze algorithmic parameters, and build a prototype coded HDR camera that demonstrates the utility of convolutional sparse HDRI coding with a custom hardware platform.*

Categories and Subject Descriptors (according to ACM CCS): I.4.1 [Image Processing and Computer Vision]: Digitization and Image Capture—

## 1. Introduction

One of the fundamental characteristic of a sensor is its *dynamic range*: the interplay of full-well capacity, noise, and analog to digital conversion. The ability to simultaneously record and distinguish very low signals alongside extremely bright scene parts is critical for many applications in scientific imaging, microscopy, and also consumer photography. Unfortunately, the hardware capabilities of available image sensors are insufficient to capture the wide range of intensities observed in natural scenes. This has motivated researchers to develop computational imaging techniques to overcome the dynamic range constraints of sensor hardware by co-designing image capture mechanisms and post-processing algorithms.

Today, high dynamic range (HDR) photography is well-established and usually done via one of three general approaches: sequentially capturing and subsequently fusing multiple different exposures (e.g., [MP95, DM97]), capturing different exposures simultaneously with multiple sensors (e.g., [TKTS11]), or coding per-pixel or per-scanline exposures within a single image with appropriate reconstruction algorithms [NM00, NN02, GHMN10, WIH10, KGBU13, HST*14, ZSFC*15]. Whereas sequential image capture is easily afforded by existing cameras, this method makes it challenging to capture dynamic scenes and usually requires additional motion stabilization and de-ghosting techniques. Multi-sensor solutions are elegant, but more expensive and they require precise calibration. In this paper, we advocate for coded pixel exposure techniques and propose a new reconstruction algorithm for this class of computational cameras. Our approach builds on recent advances in convolutional sparse coding and reconstruction techniques. We show that a naïve application of traditional, patch-based (i.e. non-convolutional) sparse reconstruction techniques [LBRN07, CENR10] struggles to deliver high image quality for high contrast scenes. We make the key observation that convolutional sparse coding (CSC) (e.g., [KSIB*10]), is particularly well-suited for the type of high-contrast signals present in HDR images. Therefore, we pose the HDR recovery problem as convolutional sparse coding problem and derive necessary formulations to solve it efficiently. We make the following contributions:

- We introduce convolutional sparse coding (CSC) for high dynamic range image reconstruction.
- We propose forward and inverse methods that are tailored to recovering a high-contrast (HDR) image from a single, coded exposure photograph.
- We demonstrate improved image quality over other existing approaches and over a naïve application of sparse reconstruction techniques to HDRI. We also evaluate algorithmic parameters, analyze different exposure coding schemes, and interpret HDR image features.
- We build a prototype coded exposure camera and demonstrate the utility of our algorithm using data captured with this prototype.

## 2. Related Work

One of the most common techniques to compute HDR images is exposure bracketing. This technique, also known as multi-bracketing, merges several LDR images of the scene taken with different bracketing exposures, into the final HDR image [MP95, DM97]. One of the main drawbacks of this technique is that, if either the camera or some scene elements move during the extended capture process, ghosting artifacts appear. There have been many algorithms designed to remove these artifacts by means of alignment and de-ghosting [SS12]. Some recent works include the use of optical flow [ZBW11], patch-based reconstruction [GGC*09, SKY*12], or modeling the noise distribution of color values [GKTT13]. The problem is further aggravated for HDR video (e.g., [KSB*13, MG11]): on the one hand, optical flow solutions fail in the presence of complex motion, on the other hand patch-based methods lack built-in temporal coherence. In contrast, the proposed convolutional sparse coding approach can produce an HDR image from a single shot, thus removing the need for alignment, motion estimation or, in general, any de-ghosting strategy.

Other works rely on multiple cameras [SBB14, BRG*14], enhanced sensor control electronics performed in simulation [PZJ13], or otherwise highly modified hardware designs [MRK*13, ZSFC*15]. For instance, Tocci et al. [TKTS11] and Kronander et al. [KGBU13] achieve single-shot HDR by acquiring several LDR images with different sensors using a beam splitter. Our method uses an off-the-shelf camera with a simple mask on the sensor or using a per-pixel coding exposure, which greatly reduces complexity, size and overall cost.

Previously proposed single-shot approaches rely on exposures that vary per image scanline, for example implemented with coded electronic shutters [CKL14], or sensors which allow different gain settings simultaneously for alternating pixel rows [GHMN10, HKU14, HST*14]. In all of these cases, an image is reconstructed using sophisticated interpolation methods, and often relies on additional image priors. These methods present a trade-off between the dynamic range that can be recovered with only two different exposures, and the quality of the final reconstruction, determined by how far apart the exposures are chosen. Other spatially-varying gain methods aim at capturing increased dynamic range from a single image, using a per-pixel coded exposures. Nayar and colleagues [NM00, NN02] place a mask of spatially varying neutral density filters on the sensor, effectively coding different exposures for adjacent pixels according to the optical pattern of the mask. However, this method is limited by interpolation artifacts and aliasing resulting from the regular pattern of the mask. The work by Aguerrebere and colleagues [AAD*14] leverages recent advances in solving inverse problems [YSM12] together with a spatially-varying mask, but still relies on a complex MAP Expectation-Maximization optimization framework which can lead to artifacts in scenes of high dynamic range.

In this paper, we propose a sparse reconstruction framework that takes advantage of the compressibility of visual information to reconstruct a high dynamic range image from asingle shot with pixel-coded exposure. Sparse reconstruction has been used before in the context of rendering [SD11], and image reconstruction and acquisition [SD09b, MKU15], including high-speed video [LGH*13, SGM15], dual photography [SCG*05, SD09a] and light transport acquisition [PML*09], light field capture [MWBR13], hyperspectral imaging [LLWD14, JCK16], or even extended dynamic range imaging using a Fourier basis [SBN*12, SKZ*13]. However, we do not rely on a conventional, patch-based learning and reconstruction method as most of these works do because it has certain limitations for the recovery of HDR images. Instead, we propose a novel formulation based on convolutional sparse coding (CSC). CSC has been used for learning hierarchical image representations [KSIB*10, ZTF11, CPS*13] and to solve transient imaging problems [HXK*14, HDL*14]. We build on the basic idea of convolutional sparse coding and make it practical for coded, single-shot HDR image acquisition.

## 3. CSC framework for HDR reconstruction

In this section, we offer a brief review of sparse coding techniques and introduce a new formulation of convolutional sparse coding tailored to the problem of high dynamic image reconstruction from a single image with spatially-varying pixel exposures.

### 3.1. Review of sparse coding and reconstruction

The traditional problem faced in sparse reconstruction is that of solving an underdetermined system of linear equations $\mathbf{y} = \mathbf{\Phi}\boldsymbol{\alpha}$ in which $\boldsymbol{\alpha} \in \mathbb{R}^n$ is the signal we are interested in, $\mathbf{y} \in \mathbb{R}^m$ is the signal we actually can measure, and $\mathbf{\Phi} \in \mathbb{R}^{m \times n}$ is the sensing matrix, such that $m < n$.

Solving the sparse reconstruction problem relies on the assumption that the signal is sufficiently compressible in some basis or dictionary $\mathbf{\Lambda} \in \mathbb{R}^{n \times l}$. This implies that $\boldsymbol{\alpha} = \mathbf{\Lambda}\mathbf{s}$, with most coefficients of $\mathbf{s} \in \mathbb{R}^l$ being zero or close to zero. This dictionary is often learned from a training set representative of the images of interest[†] [AEB06, MBPS09]. We can then recover $\boldsymbol{\alpha}$ under certain conditions by solving the following minimization problem [Ela10]:

$$\min_{\mathbf{s}} \|\mathbf{s}\|_1 \quad \text{subject to} \quad \|\mathbf{y} - \mathbf{\Phi}\mathbf{\Lambda}\mathbf{s}\|_2 \leq \varepsilon \qquad (1)$$

where $\varepsilon$ represents uncertainties in the measurements, such as sensor noise. This minimization is solved in a patch-based manner, that is the image is divided into a series of overlapping patches and each patch is reconstructed individually using Eq. 1. All the reconstructed patches are subsequently merged, for example by computing a per-pixel average, to yield the final result.

A drawback of dictionary-based sparse coding approaches is that important spatial structures of the signal of interest can be lost due to the subdivision into mutually-independent patches. Further, patches (atoms) of the dictionaries learned with this approach are often redundant and contain shifted versions of the same features. This can be seen in Figure 2 (left), which shows sample atoms of a dictionary learned from HDR images. Moreover, as we show in Section 4.2 and Figure 5, due to the nature of the mathematical formulation (a linear combination of learned patches), these patch-based approaches can fail to adequately represent high-frequency, high-contrast image features, which are particularly important in HDR images.

An alternative to patch-based approaches is CSC, which instead is based on an image decomposition into spatially-invariant convolutional features, as explained in the following. Compared to the atoms of a dictionary, the learned filters of our CSC scheme (Figure 2 (right)) show a much richer variance (e.g., they span a larger range of orientations), which leads to better reconstructions.

Convolutional sparse coding models the signal of interest $\boldsymbol{\alpha} \in \mathbb{R}^n$ as a sum of sparsely-distributed convolutional features [HHW15], that is $\boldsymbol{\alpha}$ is modeled as:

$$\boldsymbol{\alpha} = \sum_{k=1}^{K} \mathbf{d}_k * \mathbf{z}_k, \qquad (2)$$

---

[†] Alternatively, well-explored sparsity bases, such as the DCT or wavelets, could be used.

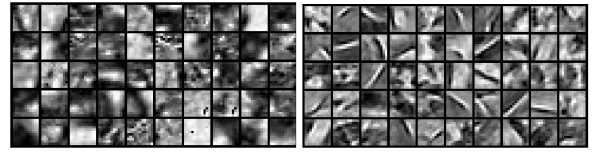

**Figure 2:** *Left: sample atoms of learned dictionary trained on HDR images (patches are tonemapped for display). Right: sample filters learned with a convolutional sparse coding framework. The convolutional filter bank shows less redundancy, crisper features, and a larger range of feature orientations.*

In this case, the dictionary is a convolutional filter bank formed by filters $\mathbf{d}_k$ of fixed spatial support $\sqrt{p} \times \sqrt{p}$, while $\mathbf{z}_k$ are sparse feature maps of size $\sqrt{n} \times \sqrt{n}$.

Consequently, the signal recovery can be performed by solving

$$\underset{\mathbf{d}, \mathbf{z}}{\operatorname{argmin}} \frac{1}{2} \|\mathbf{x} - \sum_{k=1}^{K} \mathbf{d}_k * \mathbf{z}_k\|_2^2 + \beta \sum_{k=1}^{K} \|\mathbf{z}_k\|_1$$
$$\text{subject to} \quad \|\mathbf{d}_k\|_2^2 \leq 1 \quad \forall k \in \{1, \dots, K\}. \qquad (3)$$

Heide and colleagues [HHW15] generalized this formulation to be able to handle incomplete data, as modeled by the general linear operator $\mathbf{M}$:

$$\underset{\mathbf{d}, \mathbf{z}}{\operatorname{argmin}} \frac{1}{2} \|\mathbf{x} - \mathbf{M} \sum_{k=1}^{K} \mathbf{d}_k * \mathbf{z}_k\|_2^2 + \beta \sum_{k=1}^{K} \|\mathbf{z}_k\|_1$$
$$\text{subject to} \quad \|\mathbf{d}_k\|_2^2 \leq 1 \quad \forall k \in \{1, \dots, K\}. \qquad (4)$$

They also proposed a technique for efficiently solving this problem via splitting of the objective function.

### 3.2. HDR image formation model

Based on the film reciprocity equation [DM97], we can describe the image formation model at the sensor as:

$$\mathbf{y} = f(\mathbf{p} * \Delta t \mathbf{L}) \qquad (5)$$

where $\mathbf{y} \in \mathbb{R}^n$ is the vectorized image captured at the sensor, $\Delta t$ is the exposure time, $\mathbf{L} \in \mathbb{R}^n$ represents radiance values, and the function $f$ models the camera response. The convolution by $\mathbf{p}$ is modeling the effect of the point spread function (PSF) of the optical system, which can also be expressed as a multiplication by a convolution matrix $\mathbf{P}$. Note that we use radiance $\mathbf{L}$ instead of irradiance since almost all modern cameras provide a nearly constant mapping between both magnitudes, compensating for angular effects [DM97, KMH95]. We optically modulate the light arriving at each pixel by placing a coded transmissivity mask $\mathbf{\Omega}$ on the sensor or by applying a spatially-coded exposure readout. This can be formulated as

$$\mathbf{y} = f(\mathbf{\Omega}\mathbf{P}\Delta t \mathbf{L}) \qquad (6)$$

where $\mathbf{\Omega} \in \mathbb{R}^{n \times n}$ is a diagonal matrix containing the modulation code of the mask. For RAW images, we can assume a linear response of the digital sensor with respect to irradiance for all non-saturated pixels [LMS*13]. Thus, we can rewrite Equation 6 as $\mathbf{y} = \zeta \mathbf{\Omega PL}$, where $\zeta$ is a scale factor modeling the linear response of the sensor and the influence of exposure time $\Delta t$. This scaling factor (and thus absolute radiance values) could be recovered by imaging a calibrated light source and scaling all radiance values accordingly. In our context, we aim at obtaining relative radiance values, therefore we can remove $\zeta$ and rewrite Equation 7 in normalized form as:

$$\mathbf{y} = \mathbf{\Omega PL}^* \qquad (7)$$

where $\mathbf{L}^*$ represents relative radiance values. The mask $\mathbf{\Omega}$ will ensure that pixels are sampled with effectively different exposure values, so that in all image regions at least some of the pixels properly sample the dynamic range. The sparse reconstruction step described next will be in charge of obtaining the radiance values from these differently sampled pixels.

### 3.3. Convolutional sparse HDRI coding

Equation 4 allows for the recovery of contrast-normalized images in which part of the data is missing or unreliable, as given by matrix $\mathbf{M}$. In the case of HDR reconstruction, however, our captured image $\mathbf{y}$—as given by Eq. 7—does not only have missing or unreliable data, but also differently exposed pixels due to matrix $\mathbf{\Omega}$. In the case of HDR imaging the unreliable data $\mathbf{M}$ corresponds to both saturated and noisy pixels. Incorporating the varying exposures $\mathbf{\Omega}$ we pose the convolutional reconstruction of radiance values as:

$$\underset{\mathbf{z}}{\operatorname{argmin}} \ \frac{1}{2} \|\mathbf{y} - \mathbf{\Omega MP} \sum_{k=1}^{K} \mathbf{d}_k * \mathbf{z}_k\|_2^2 + \beta \sum_{k=1}^{K} \|\mathbf{z}_k\|_1 \qquad (8)$$

where $\beta$ controls the relative weight of the sparsity term. Note that, in contrast to Eqs. 3 and 4, we optimize only for $\mathbf{z}$, since we assume that we have already learned a dictionary of filters $\mathbf{d}$.

The dictionary of filters $\mathbf{d}$ is learned using Eq. 4, and some of the learned filters are shown in Figure 2 (right). We learn the filters from a set of LDR images, after performing a local contrast normalization on these images. This amounts to learning from whitened data (normalized sigma and mean). As a consequence of this normalization, the formulation cannot be used directly in a generative model: While the correct scaling for recovery can be obtained during the optimization by finding the correct values in the sparse maps $\mathbf{z}$, the offset cannot. To solve this, we introduce an offset term $\mathbf{o}$ that we jointly estimate with the sparse feature maps. The smoothness is ensured by a quadratic smoothness constraint, leading to:

$$\underset{\mathbf{z}}{\operatorname{argmin}} \ \frac{1}{2} \|\mathbf{y} - \mathbf{\Omega MP}(\sum_{k=1}^{K} \mathbf{d}_k * \mathbf{z}_k + \mathbf{o})\|_2^2 + \beta \sum_{k=1}^{K} \|\mathbf{z}_k\|_1 + \lambda_s \|\nabla \mathbf{o}\|_2^2 \qquad (9)$$

Thanks to this normalization, the filters generalize to different means and scales—which are obtained during the optimization—, and they are independent of dynamic range. We additionally observe that the learned filters have fewer data-specific features and are more general this way, and the learning converges in fewer iterations. Specific implementation details on the filter dictionary learning are given in Section 5.

We can elegantly fit this additional offset in the proposed optimization framework by expressing it as the convolution $\mathbf{o} = \mathbf{d}_{K+1} * \mathbf{z}_{K+1}$, where $\mathbf{d}_{K+1}$ is a Dirac delta, and Equation 9 thus becomes:

$$\underset{\mathbf{z}}{\operatorname{argmin}} \ \frac{1}{2} \|\mathbf{y} - \mathbf{\Omega M} \sum_{k=1}^{K+1} \mathbf{Pd}_k * \mathbf{z}_k\|_2^2 + \beta \sum_{k=1}^{K} \|\mathbf{z}_k\|_1 + \lambda_s \|\nabla \mathbf{z}_{K+1}\|_2^2 \qquad (10)$$

where $\lambda_s$ controls the relative weight of the smoothness term. Note that only smoothness, and not sparsity, is enforced for this $\mathbf{z}_{K+1}$.

Finally, if we rewrite Equation 10 by substituting $\hat{\mathbf{M}} = \mathbf{\Omega M}$ and $\hat{\mathbf{d}}_k = \mathbf{Pd}_k$, our problem can be written as the CSC problem shown in Equation 4, with the exception of the quadratic smoothness term:

$$\underset{\mathbf{z}}{\operatorname{argmin}} \ \frac{1}{2} \|\mathbf{y} - \hat{\mathbf{M}} \sum_{k=1}^{K+1} \hat{\mathbf{d}}_k * \mathbf{z}_k\|_2^2 + \beta \sum_{k=1}^{K} \|\mathbf{z}_k\|_1 + \lambda_s \|\nabla \mathbf{z}_{K+1}\|_2^2 \qquad (11)$$

We solve this problem using a modification of the ADMM algorithm [BPC*11]. To do so, we need to reformulate Equation 11 to express the first two terms as a sum of functions, in the following form:

$$\underset{\mathbf{z}}{\operatorname{argmin}} \ \sum_{i=1}^{I} f_i(\mathbf{K}_i \mathbf{z}) + \lambda_s \|\nabla \mathbf{z}_{K+1}\|_2^2, \qquad (12)$$

For more details on this transformation please refer to [HHW15, Sec. 2.1 and 2.2]. Once this is done, the modified ADMM algorithm to solve for $\mathbf{z}$ in our case is shown in Algorithm 1. The update in line 2 of the algorithm is solved in the spectral domain, and thus the additional smooth constraint does not increase the computational cost significantly w.r.t. the original formulation [HHW15]. Also, the filter size does not matter in our case, since we are performing the filter inversion in the frequency domain. This would not be computationally efficient with traditional CSC methods such as that of Szlam et al. [SKL10]. Finally, $\mathbf{prox}_{\phi}$ refers to the proximal operator of a function $\phi$ as described in Parikh and Boyd's work [PB14].

---

**Algorithm 1** ADMM for HDR recovery

1: **for** $k = 1$ to $V$ **do**
2: $\quad \mathbf{y}^{k+1} = \underset{\mathbf{y}}{\operatorname{argmin}} \ \|\mathbf{Ky} - \mathbf{z} + \lambda^k\|_2^2 + \lambda_s \|\nabla \mathbf{z}_{K+1}\|_2^2$
3: $\quad \mathbf{z}_i^{k+1} = \mathbf{prox}_{\frac{f_i}{\rho}}(\mathbf{K}_i \mathbf{y}_i^{k+1} + \lambda_i^k)$
4: $\quad \lambda^{k+1} = \lambda^k + (\mathbf{Ky}^{k+1} - \mathbf{z}^{k+1})$
5: **end for**

---

### 4. Analyzing convolutional sparse HDRI coding

In this section, we provide an analysis of the proposed framework, including choice of coded exposure patterns and algorithmic parameters. We also show advantages of this formulation over traditional, patch-based sparse reconstruction for HDR capture.

## 4.1. Design of coded exposure patterns

There are several factors to take into account when designing the optical mask $\Omega$. First, it needs to have a high light throughput, to avoid noise and reduce required exposure time; second, its per-pixel transmissivity values $e_i$ should cover a wide range of exposures (that is, $e_{max}/e_{min}$ should be large); and third, it should facilitate practical implementation. We tested several configurations for the mask over a set of seven different images; in particular, these configurations were: binary, Gaussian, uniform, uniform with four fixed exposures, fixed pattern with four exposures, and interleaved exposure. In the following we detail the formulation for each mask, the motivation behind its testing, and its performance.

We initially tested and compared the performance three optical masks: a binary mask, a mask where exposure values are drawn from a Gaussian distribution ($\Omega_G = \{e_i \; ; \; e_i \sim \mathcal{N}(0.6, 0.1)\}$), and a mask obtained by drawing values from a uniform distribution ($\Omega_U = \{e_i \; ; \; e_i \sim \mathcal{U}(0, 1)\}$). The reconstruction results are shown in Figure 3. The binary mask is limited when modulating the incoming light, and, as a result, is very limited in terms of the recovered dynamic range; large saturated areas, for instance, will be impossible to recover since all the pixels will be degraded due to the binary sampling. Both the uniform and the Gaussian masks yield good results, and choosing between them represents a trade-off between transmissivity and dynamic range. The Gaussian mask offers better light throughput, but a more limited recoverable dynamic range: most of the values of the Gaussian distribution will be close to the mean, with few very low values. As a result, large bright areas (such as in Figure 3, around the sun) may still remain saturated. A uniform mask allows recovery of a larger dynamic range because it more uniformly samples the range of exposures, minimizing the risk of large under- or over-exposed areas even in scenes of very high dynamic ranges.

While a uniform mask works well in practice, for a practical hardware implementation having a low number of discrete exposure values is beneficial. We therefore compare the uniform mask $\Omega_U$ with a uniform 4-exposure mask $\Omega_F$, that is one in which each pixel randomly takes one of four exposure values $\{e_1 .. e_4\}$. We choose the exposure values such that the ratio $e_{max}/e_{min}$ covers 6 f-stops, i.e., $e_4/e_1 = 2^6$; this, with the dynamic range of 1000:1 that a standard CMOS sensor has [EG02], allows us to recover up to 16 stops in dynamic range. Figure 4 shows the quality of the resulting reconstruction for $\Omega_U$ and $\Omega_F$, which can be seen to be very similar in both. Thus, $\Omega_F$ allows us to recover a very similar range to the uniform one, without artifacts, and has an easier implementation. Consequently, in the remainder of the paper, we opt for a uniform, 4-exposure pattern ($\Omega = \Omega_F$), since it offers the best trade-off between quality of the results—in terms of recovered dynamic range and absence of artifacts—, and ease of implementation in hardware. The exception to this is our hardware prototype (Section 5.1): since it exhibits significant light loss (mainly due to the LCoS and the beamsplitter) we do use a Gaussian mask to minimize the impact of the reduced light throughput. However, future chip designs with built-in per-pixel exposure will overcome this prototype's limitations; taking this into account the best option among the configurations we tested is $\Omega_F$.

Additionally, to highlight the versatility of our reconstruction framework, we tested two additional exposure patterns which have been used before in the context of HDR imaging. Their results are also shown in Figure 4 (in the two rightmost images). In particular, we show a reconstruction result for a *fixed pattern* $\Omega_P$, using four exposures (that is, the mask shows a repeating, fixed $2 \times 2$ pattern), and a result for an interleaved exposure pattern. The former has been proposed before for HDR imaging, but with the reconstruction done by means of interpolation [NM00], which can lead to aliasing effects. The latter is inspired by the *Magic Lantern* software package, which offers a firmware upgrade to capture an interleaved exposure consisting of alternating rows with two different exposures $\Omega_I$ for some off-the-shelf cameras. Our framework allows for a plausible result even with these exposure patterns.

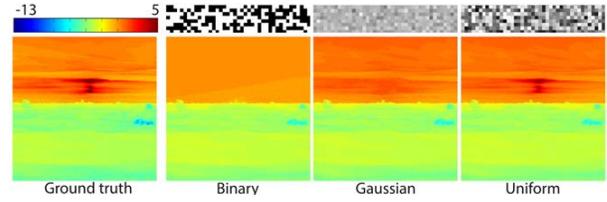

**Figure 3:** *HDR images in false color (color scale shows f-stops) showing (from left to right): ground truth radiance, radiance recovered using a binary mask $\Omega_B$ as optical code, a Gaussian mask $\Omega_G$, and a uniform mask $\Omega_U$ (more details in the text). The first two masks clearly fall short when recovering dynamic range, while the uniform one offers results very close to the original. The tonemapped ground truth image can be seen in Figure 4, left.*

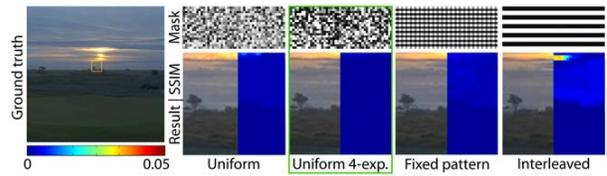

**Figure 4:** *Left: a tonemapped HDR ground truth image. Right: quality of different optical masks when attempting to recover the ground truth scene radiance. From left to right: a uniform mask $\Omega_U$, a 4-exposure mask $\Omega_F$, an interleaved mask $\Omega_I$, and a fixed pattern mask $\Omega_P$. For each one, the left part shows the reconstructed image, and the right part the error with respect to the ground truth displayed as $(1 - SSIM)$ [WBSS04]. The top row shows a sample region of the corresponding mask. We choose $\Omega_F$ for its ability to faithfully recover a wide dynamic range and its ease of implementation. Please refer to text for more details.*

## 4.2. Advantage of CSC HDRI over patch-based approaches

Patch-based sparse reconstruction approaches have been widely used in computational imaging problems [LGH\*13, MWBR13, LLW14]. In this section, we illustrate and explain how directly applying such approaches to the problem of HDR reconstruction

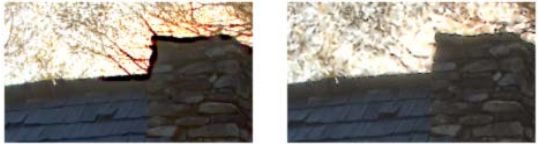

**Figure 5:** *Detail of an HDR image reconstructed using a patch-based sparse reconstruction approach (left) and our convolutional sparse coding framework (right). The former is unable to recover very high-contrast sharp edges, while the latter offers good results in this case. The images are tonemapped for display using [MDK08].*

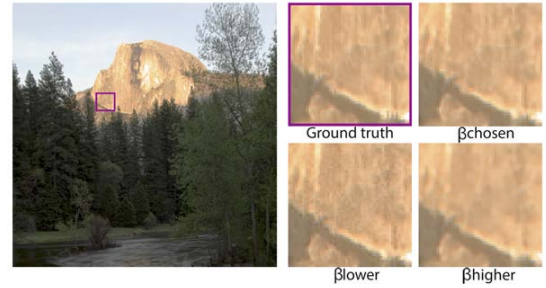

**Figure 6:** *Reconstructed HDR image (tonemapped for display) showing the effect of β, the relative weight of the sparsity term, in the optimization. Please refer to the text for details.*

from a single, exposure-coded image would produce undesired results in a number of cases.

We have already seen how the filters in our framework show a richer variance (less redundancy and a larger range of orientations) compared to traditional atoms in a learned dictionary (Figure 2). Consequently, CSC dictionaries are more descriptive and better capture the essence of the signals (e.g., they avoid the need to have shifted versions of the patches), which results in better reconstructions. More importantly, in patch-based approaches, the signal (a given image patch that is to be reconstructed) is represented as a linear combination of dictionary patches with their associated coefficients. This is problematic when attempting to reconstruct patches which contain very large contrast edges (common in HDR images), because an extremely large number of patches with high-valued coefficients is needed to properly reconstruct the edge. This is of course not only the case with learned dictionary patches, but also if any other basis (e.g., DCT) is used. As such, this problem was also encountered in the past in HDR image compression [MKMS04, Fig. 5]. Consequently, when reconstructing HDR images with a patch-based approach the reconstruction fails in the presence of very high contrast edges, yielding artifacts as shown in Figure 5, left. CSC, in contrast, can naturally handle these large contrast edges—as shown in Figure 5 (right)—thanks to the formulation of the signal as a sum of convolutions of the filters by sparse feature maps as opposed to a linear combination of dictionary elements.

Moreover, the convolutional sparse coding framework converges significantly faster than the patch-based approach (for which we use the well-known OMP algorithm [TG07]). Specifically, in an Intel Xeon E5-1620 @3.50GHz with 16GB RAM our CSC approach is around 2.5x faster.

### 4.3. Optimization parameters

As explained in Section 3.3, β controls the relative weight of the sparsity term with respect to the data term (see Equation 11). Increasing the value of β will therefore result in a degradation of the high frequencies in the reconstructed scene, since the feature maps **z** will be too sparse to represent fine details. Decreasing β, on the contrary, will lead to an excessive relative weight of the data term, which can result in artifacts due to approximations of non-linearities of the process (such as the quantization). Fig-ure 6 shows this behavior. We choose an intermediate value of β, $\beta_{chosen} = 1.5 \cdot 10^{-5}$, which we use in all the reconstructions shown in this work.

The other relevant parameter in the optimization is the relative weight of the quadratic smoothness term, $\lambda_s$ in Equation 11; we choose $\lambda_s = 0.5 \cdot 10^{-5}$. In this case, it is important that a good estimate of the offset term $z_{K+1}$ is given as initial value to the optimization. We provide a blurred version of the captured LDR image divided by the optical mask, which yields good results and fast convergence.

## 5. Results

We show here reconstruction results using both existing HDR images[‡], and data captured with our prototype camera. All results shown have been reconstructed using our single-shot method described in this paper, with the same optical mask $\mathbf{\Omega}_F$ described in Section 4.1, consisting of four randomly sampled exposure values with $e_{max}/e_{min} = 2^6$, except where otherwise indicated. The filter bank $\mathbf{d}_k$ used for the reconstruction is learned from a collection of ten natural LDR images using the method proposed by Heide et al. [HHW15]; a representative sample of these learned filters is shown in Figure 2 (right). When choosing the training images we learn the filters from, we found our framework robust enough to provide similar results when learned from different sets of images: Learning the filter bank from a dataset of images used in the work of Heide et al. or learning from tonemapped images from Fairchild's database (on a set not used for testing) yielded reconstructions which differed in less than 0.5 dB in PSNR. The size of the filters is determined by the resolution of the training data; the filters need to be large enough so they contain useful information, yet small enough not to overfit to specific features of the training data. We find that learning $K = 100$ filters of size $11 \times 11$ pixels fulfills these conditions for our data and works well for all the images tested. All HDR results shown have been tonemapped using the same algorithm [MDK08]. We additionally compare our results to two other spatially varying exposure methods [NM00, HKU14]. For results using existing HDR images as input, we simulate the process of capturing the coded LDR image as follows: We first apply a convolution kernel **p** simulating the optical PSF of the camera, and modulate light arriving at the sensor multiplying the radiance values of the input HDR by our coded mask. We bracket these values taking into account that a typical CMOS sensor has a dynamic range of around 1000 : 1. In doing so, we assume a reasonably well-exposed LDR image, but nevertheless we simulate the metering of a camera and take into account saturation and under-exposure by placing the sensor range so that the number of saturated and under-exposed pixels is minimized. Then we normalize these bracketed values and apply a camera response function[§]. Last, we quantify the resulting values to store the LDR image which will be used as input for the reconstruction.

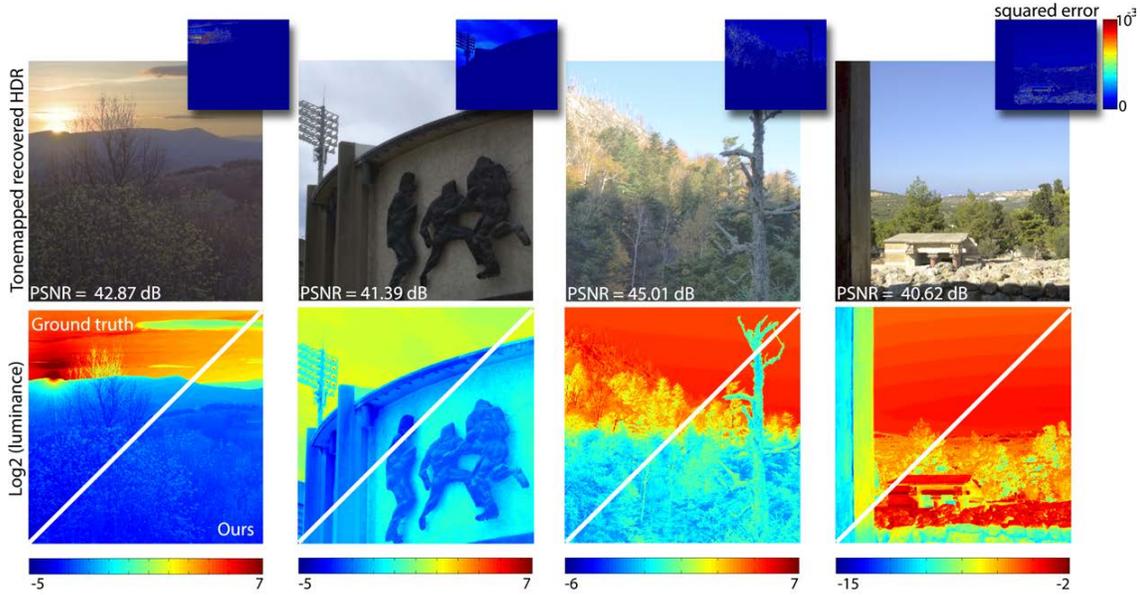

**Figure 7:** Top row: *Recovered HDR images from a single-shot coded image (tone mapped using [MDK08]), and PSNR values. The insets show the squared, per-pixel difference with respect to the ground truth luminance.* Bottom row: *False color (split) images depicting luminance of the original scene, and of our reconstructed scene; we use a base-2 logarithm to properly display the extremely large dynamic range.*

Figure 7 shows four of our reconstructed HDR images. In addition to our reconstruction (top row), we show, for each scene, a false color image of the ground truth scene and our reconstruction (bottom row, split images) to show our ability to recover the large dynamic range present in the original scene. Since we recover relative radiance, and given the large dynamic range, we plot in false color $log2$ radiance normalized to the ground truth. Further, the insets in the top row show the error, computed as the square of the per-pixel difference between ground truth and our reconstruction, scaled for visualization purposes. We also report the PSNR for each one, which is always above 40 dB. This figure shows how our method is able to recover scenes with very high dynamic range, faithfully reproducing contrast in the original scene. More reconstructed scenes can be found in Figure 8, in which we show our reconstructed HDR image (top row), normalized luminance of sample scanlines, both recovered and ground truth (middle row), and two exposures of the reconstructed scene to better show the quality of the reconstruction across the dynamic range, including the challenging case of high-contrast sharp edges (bottom row).

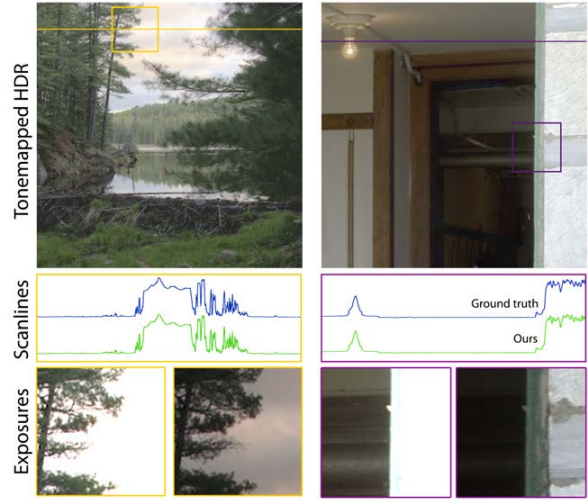

**Figure 8:** *Additional results obtained by our technique for two HDR scenes.* Top row: *tonemapped HDR image (using [MDK08].* Middle row: *Normalized luminance plots for the corresponding marked scanlines for our recovered image (green curve) and the ground truth image (blue curve).* Bottom row: *Close-up of two exposures of the corresponding highlighted regions, displaying very high-contrast edges.*

---

[§] http://www1.cs.columbia.edu/CAVE/software/softlib/dorf.php

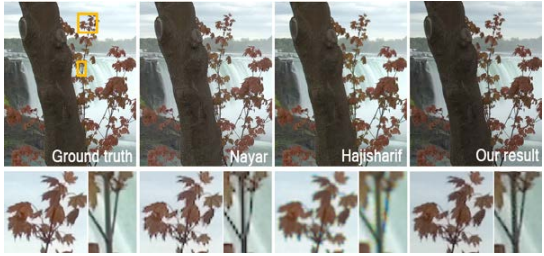

**Figure 9:** *Comparison with two representative spatially varying exposure methods [NM00, HKU14]. The inherent interpolation step in such methods leads to visible artifacts in areas of high contrast or very fine detail.*

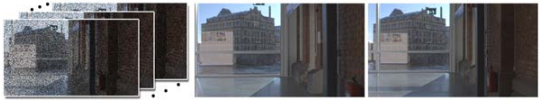

**Figure 10:** *HDR reconstruction of an animated scene. Left: Coded sensor images using our optical code $\Omega_F$. Center and right: Two example frames exhibiting temporal coherence in the reconstruction. Input video from the LiU HDR video repository (http://www.hdrv.org).*

Different from other common spatially varying exposure methods, our approach does not rely on interpolation of the captured samples to reconstruct the image. Instead, it exploits information of the structure of natural images through the learned convolutional filter bank, which greatly minimizes the presence of visible artifacts in areas of high contrast or very fine detail. We show this by explicitly comparing our results against the spatially varying exposure methods of Nayar et al. [NM00] and Hajisharif et al. [HKU14], which makes use of the *Magic Lantern* software to capture interlaced, dual-ISO images. Our method preserves edges better, minimizing the aliasing artifacts that arise from the trade-off between spatial resolution and dynamic range in Nayar's method, while Hajisharif's method has difficulties recovering thin structures, such as the small branches of the tree (Figure 9).

Our technique can be applied to the reconstruction of HDR animated scenes as well, using the same optical code for each frame. Our reconstruction framework yields a very faithful recovery of the original signal, naturally leading to temporal coherence, without the need for explicit enforcement. We show this in Figure 10, using an existing HDR video from the LiU HDR video repository[¶], and also include this video in the supplementary material. The HDR video recovering is performed frame by frame from LDR capture simulations from the aforementioned HDR video.

Finally, our framework can also be used for compression of HDR images. Traditional techniques used for compression of images can fail when applied to HDR images, due to the high-contrast sharp edges that can be present in them. Consequently, techniques have been developed to compress this type of content [MKMS04]. Our framework allows for compression of HDR images, since we can represent them with a set of sparse feature maps. We have shown in Figure 5 how for HDR content we avoid artifacts that appear when codification and reconstruction with patch-based schemes is used. Note that DCT was also proven to not work well by Mantiuk et al. [MKMS04], requiring more complex processing for compression.

### 5.1. Hardware prototype implementation

Per-pixel exposure cameras are not commercially available yet, although a per-pixel exposure patent has already been filed by Sony Corporation [Jo14]. We have built a prototype that simulates this feature to demonstrate our method with real scenes. To this end, we have implemented a capture system based on a liquid crystal on silicon (LCoS) display (Figure 11, left). This device, together with a beamsplitter and relay optics, simulates a Gaussian attenuation mask placed before the sensor. In this setup, the SLR camera lens (Canon EF-S 60 mm f/2.8 Macro USM) is focused on the LCoS, virtually placing the mask at the sensor. Our imaging lens is a Canon EF 50 mm f/1.8 II, focused at 50 cm; scenes are placed at 80 - 100 cm. The f-number of the system is f/2.8, the maximum of both lenses. Since a single pixel of the LCoS cannot be well-resolved with this setup, we treat LCoS pixels in blocks of $8 \times 8$ pixels, resulting in a mask with a resolution of $240 \times 135$. Figure 11 (right) shows results with real scenes captured with our prototype optical setup. The figure includes a close-up of the LDR coded image captured at the sensor, the final tone mapped HDR reconstruction, and several details with varying exposure levels. Our lab prototype is not artifact-free, although it demonstrates the viability of our approach. The LCoS displays some birefringence, decreased light throughput, and a severe loss of contrast, all of which degrade the LDR captured signal. Future chip designs such as the Sony patent could overcome these limitations. Nevertheless, our reconstruction does not introduce additional degradation in the results, as Figures 7 and 8 show.

Additionally, we have applied our technique to an image captured using an interlaced exposure with dual ISO 100/800 on a Canon EOS 500D camera with the *Magic Lantern* sofware. The result is shown in Figure 12.

### 6. Discussion and conclusion

**Limitations** In some cases, it is possible that the image **y** captured with the optical mask contains large saturated areas despite the presence of the mask; the low transmissivity pixels of the mask typically prevent this, but in images with extremely large dynamic range it can happen. In these cases when no information at all is captured, the recovery may have some artifacts. An example of this is shown in the inset figure with a light bulb. This light bulb is a close-up region of the scene in Figure 8 (right column). This scene has a very large dynamic range (over 17 stops), since it captures both the very dark inside of the room and the bright light bulb outside. Therefore, if the inside is to be recovered, there is a saturated area in the captured image **y**. Nevertheless, as we show in the paper, we are able to faithfully reconstruct scenes of very large dynamic range.

---

[¶] http://www.hdrv.org

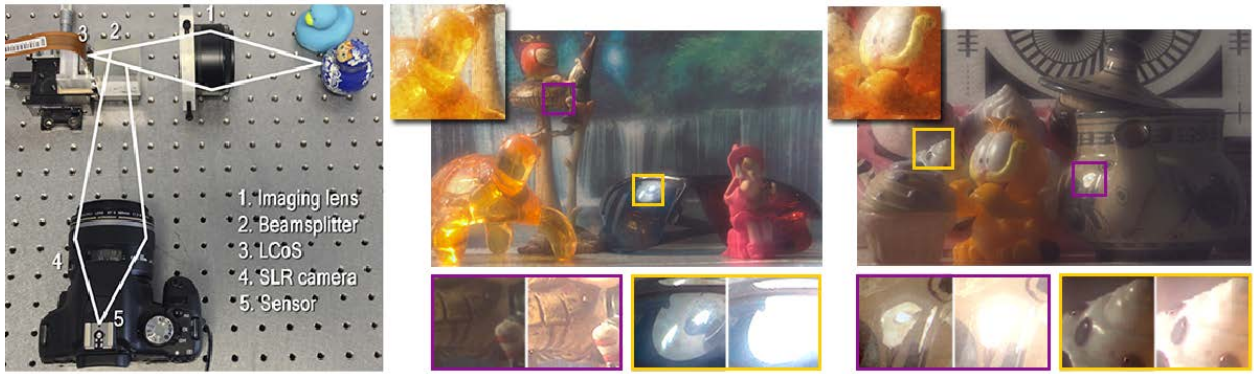

**Figure 11:** Left: *Our prototype hardware implementation. Our optical system is made up of an imaging lens, a beamsplitter, an LCoS, and an SLR camera. Objects are placed for illustration purposes only; when photographing the scene, they are placed at a distance of 80 - 100 cm from the imaging lens.* Middle and right: *Two reconstructions of real scenes. For each scene we show the tonemapped HDR reconstruction (top), two different exposures of the highlighted areas revealing the dynamic range (bottom), as well as a partial detail of the LDR coded image captured at the sensor (inset).*

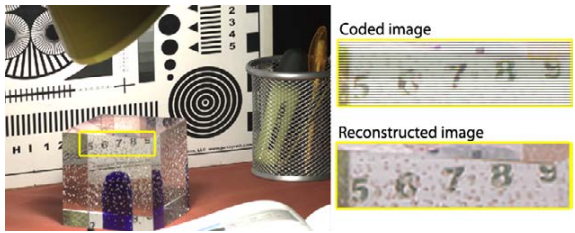

**Figure 12:** *Reconstruction of an HDR image captured with dual ISO 100/800 with a Canon EOS 500D: original scene* (left), *and close-ups of coded and reconstructed regions, the latter tonemapped using [MDK08]* (right).

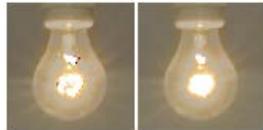

**Benefits** We have presented a framework for convolutional sparse coding of HDR images. From a single, optically coded im-age, we reconstruct dynamic range using a trained convolutional filter bank. Our approach follows a current trend in computational photography, leveraging the joint design of optical elements and processing algorithms. Once trained, the obtained filter bank can be used to reconstruct a wide variety of HDR images greatly differing from the training set. Since our reconstruction is based on a convolutional approach, it does not rely on the linear combination of patches common in sparse reconstruction methods; this greatly reduces reconstruction artifacts, in particular in high-contrast sharp edges present in HDR images. We are not limited to a restricted number of captured exposures, nor do we face the implicit trade-off between captured dynamic range and interpolation quality that other methods based on spatially-varying exposures face. In comparison to other CSC approaches, the algorithm we base our formulation on has demonstrated (see [HHW15, Sec. 3]) that it has a lower complexity and better convergence than previously proposed methods for CSC [ZKTF10, BEL13, BL14], benefits which directly carry over to our method.

As an additional advantage, our framework naturally accounts for the optical PSF of the system, since we incorporate it in our model (**P** in Equation 10). Moreover, it can be easily extended to perform demosaicking, by properly designing matrix **M** in Equation 10, which models missing pixels. Last, we have not only built a physical prototype, but have also shown how our approach can yield good results with off-the-shelf consumer hardware that captures interleaved exposures using the Magic Lantern software.

**Future work** The development of patents like Sony's per-pixel, double-exposure method will progressively introduce varying exposure and optically modulated systems, thus allowing for increased capabilities of commercial cameras. Our optimization could incorporate explicit modeling of image noise to perform denoising in particularly noisy images. Finally, an exciting avenue of future work lies at the convergence between acquisition and display technologies, for the full plenoptic function and taking perceptual considerations into account [MWDG13]; compressive sensing and sparse coding techniques may be able to handle the high dimensionality of this challenging problem.

## 7. Acknowledgements

The authors would like to thank Karol Myszkowski, as well as Jose Echevarria and Adrian Jarabo, for fruitful insights and discussion. We would also like to thank Saghi Hajisharif and Jonas Unger, for sharing their results and for their assistance with them; Nicolas Landa for preliminary testing of traditional compressive sensing on HDR; and Maria Angeles Losada and the Photonic Technologies Group at Universidad de Zaragoza for their optical instrumentation. Ana Serrano was supported by an FPI grant from the Spanish Ministry of Economy and Competitivity (project Lightslice). Felix Heide was supported by a Four-year Fellowship from the University of British Columbia. Diego Gutierrez would like to acknowledge support from the BBVA Foundation and project Lightslice. Gordon Wetzstein was supported by a Terman Faculty Fellowship and by the Intel Strategic Research Alliance on Compressive Sensing. Belen Masia was partially supported by the Max Planck Center on Visual Computing and Communication.